\documentclass[letterpaper, 10 pt, conference]{ieeeconf}  

\IEEEoverridecommandlockouts                              

\usepackage{stfloats}
\usepackage{subcaption}
\usepackage{mathtools}
\usepackage{graphics} 
\usepackage{epsfig} 
\usepackage{mathptmx} 
\usepackage{times} 
\usepackage{amsmath} 
\usepackage{amssymb}  
\usepackage{multicol}
\usepackage{graphicx}
\usepackage{epstopdf}
\usepackage{pdfpages}
\usepackage{cite}
\usepackage{amsbsy}
\usepackage{bm}
\usepackage{accents}
\newcommand*{\dt}[1]{%
  \accentset{\mbox{\bfseries .}}{#1}}
\newcommand*{\ddt}[1]{%
  \accentset{\mbox{\bfseries .\hspace{-0.25ex}.}}{#1}}
\usepackage{cite}
\usepackage{booktabs}
\pdfoutput=1
\usepackage{hyperref}
\hypersetup{
    colorlinks=true,
    linkcolor=black,
    citecolor=black,
    filecolor=black,
    urlcolor=black,
}

\def\horizontaldistance{\kern2pt}
\def\verticaldistance{5pt}

\makeatletter
\newcommand*{\rom}[1]{\expandafter\@slowromancap\romannumeral #1@}
\makeatother
\title{\LARGE \bf
Push Recovery of a Humanoid Robot Based on Model Predictive Control and Capture Point
}

\author{Milad Shafiee-Ashtiani,$^{1}$ Aghil Yousefi-Koma,$^{1}$ Masoud Shariat-Panahi,$^{2}$  and Majid Khadiv$^{3}$ 
\thanks{$^{1}$Center of Advanced Systems and Technologies (CAST)
School of Mechanical Engineering, College of Engineering, University of Tehran, Tehran, Iran.
       ( {\tt\small shafiee.a@ut.ac.ir}) ( {\tt\small aykoma@ut.ac.ir})}%
\thanks{$^{2}$School of Mechanical Engineering, College of Engineering, University of Tehran, Tehran, Iran.
       ( {\tt\small mshariatp@ut.ac.ir})}%
\thanks{$^{3}$Department of Mechanical Engineering, K. N. Toosi University of Technology, Tehran, Iran
       ( {\tt\small mkhadiv@mail.kntu.ac.ir})}%
}

\begin{document}

\maketitle
\thispagestyle{empty}
\pagestyle{empty}

\begin{abstract}
The three bio-inspired strategies that have been used for balance recovery of biped robots are the ankle, hip and stepping Strategies. However, there are several cases for a biped robot where stepping is not possible, e. g. when the available contact surfaces are limited. In this situation, the balance recovery by modulating the angular momentum of the upper body (Hip-strategy) or the Zero Moment Point (ZMP) (Ankle strategy) is essential. In this paper, a single Model Predictive Control (MPC) scheme is employed for controlling the Capture Point (CP) to a desired position by modulating both the ZMP and the Centroidal Moment Pivot (CMP). The goal of the proposed controller is to control the CP, employing the CMP when the CP is out of the support polygon, and/or the ZMP when the CP is inside the support polygon. The proposed algorithm is implemented on an abstract model of the SURENA III humanoid robot. Obtained results show the effectiveness of the proposed approach in the presence of severe pushes, even when the support polygon is shrunken to a point or a line.
\end{abstract}

\section{INTRODUCTION}
The main destination of humanoid robots research is realizing a robot that is able to work in real environments. Because of unstable nature of the biped robots, the ability of recovering from unexpected external disturbances is essential.
In recent years, several attempts have been made by researchers to generate robust locomotion of biped robots (\cite{kajita2003biped,wieber2006trajectory,pratt2006capture,herdt2010online,stephens2010push,aftab2012ankle,koolen2012capturability}). A common criterion for ensuring dynamic balance during walking is to maintain the Zero Moment Point (ZMP) or the Center of Pressure (CoP) within the support polygon of the contact points. 
The main approaches that have been used for balancing and walking of humanoid robots in the presence of disturbances are based on the Model Predictive Control (MPC) or controlling the Capture Point (CP) \cite{kajita2003biped,wieber2006trajectory,pratt2006capture,herdt2010online,stephens2010push,aftab2012ankle,koolen2012capturability,englsberger2015three,krause2012stabilization,yun2011momentum}.

\begin{figure}[]
\centering
      \mbox{\parbox{2.5in}{
 \includegraphics[scale=0.95, trim ={2.5cm 18.52cm 11.2cm 2.5cm},clip]{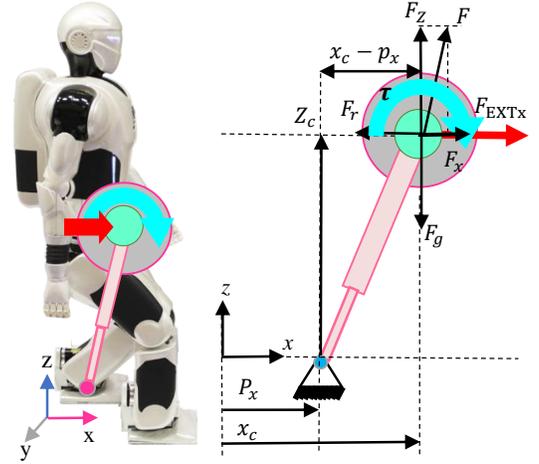}
}}
\caption{Abstract model of SURENA \rom{3} humanoid robot}
      \label{fig1}
    \end{figure}

Kajita et al.\cite{kajita2003biped} introduced preview control of ZMP and paved a way for robust walking pattern generation. This method was expressed more generally as an MPC problem by Wieber et al. \cite{wieber2006trajectory}. To increase the robustness of the gaits, the MPC formulation in \cite{wieber2006trajectory} has been modified to adapt the step locations \cite{herdt2010online,stephens2010push}. However, the upper-body angular momentum has not been employed in these works. As a result, Aftab et. al \cite{aftab2012ankle} proposed a single MPC that uses all the ankle, hip, and stepping strategies for balance recovery of humanoid robots.  In all of these works, the CoM has been considered as the state of the system. However, relating the problem in this way constrains both divergent and convergent components of motion  \cite{englsberger2015three}.

Pratt et al. \cite{pratt2006capture,koolen2012capturability} introduced the CP by splitting the Center of Mass (CoM) dynamics into stable and unstable components. The state variable related to the unstable part of the CoM dynamics has been named the Capture Point (CP). The CP specifies when and where a humanoid must step to in order to maintain balance, however it requires a controller for stabilizing unstable nature of dynamic of the CP. To this end, Englsberger et al. \cite{englsberger2015three,krause2012stabilization} developed a controller for CP tracking without using the effect of upper-body angular momentum (CMP modulating) and by guiding the CP only by CoP modulation.
The effect of upper-body angular momentum plays a key role for balance recovery especially in the situation that stepping is not possible or contact surface is small \cite{yun2011momentum,kiemel2012balance,wiedebach2016walking}.  

In this paper, in order to utilize the usefulness of the two mentioned approaches, the CP concept is used in an MPC. To do so, an effective MPC scheme is developed for push recovery by manipulating the CoP when the CP is within the support polygon, and employing the CMP modulation when the CP is out of the support polygon. The main goal of this controller is to maintain the CP, CMP and CoP on the center of support polygon. The proposed algorithm is capable of dealing with severe pushes while the contact surface is a line or a point. The remainder of this paper is organized as follows. The CoM dynamics, and the CP formulations are reviewed in Sec  \rom{2}. The proposed push recovery controller is presented in Sec  \rom{3}. In  \rom{4}, the obtained simulations results are presented and discussed. Finally, Section  \rom{5} concludes the findings.

\section{Center of Mass Dynamics}
\subsection{Linear Inverted Pendulum}

Using the full nonlinear dynamics of a humanoid robot for gait planning makes the corresponding optimization problem non-convex \cite{khadiv2015optimal}. However, the dynamics of a biped robot can be approximated by the Linear Inverted Pendulum Model (LIPM) \cite{kajita20013d}. This model is a good dynamic approximation of a biped robot, particularly for the standing posture. The LIPM uses the following assumptions \cite{kajita20013d}:

\begin{itemize}
\item The rate of change of  angular momentum is zero,
\item The CoM height remains constant
\end{itemize}

Based on the mentioned assumptions and Fig.\ref{fig1}, the equation of Motion of the LIPM can be expressed as follows:

\begin{equation}
\ddot x_{c}= \omega_{n}^{2}( x_{c}-p_{x}) 
 \label{eq:2}
\end{equation}
where $m$ is the robot mass, the CoM position is given by $P_c=[x_c,y_c,z_c]^T$, $ P_{zmp}=[p_x, p_y, 0]^T$ is the position of the ZMP and $\omega_{n}=\sqrt{(g/z_c )}$ is the natural frequency of the LIPM. The Ground Reaction Force (GRF) intersects with the CoM because the base joint of the pendulum is torque-free and the rate of change of angular momentum is zero. As shown in Fig.\ref{fig1}, $F_z$ is the vertical component of the GRF. It compensates the gravitational force $F_g$ acting on the CoM. The inertial force $ F_r=m\ddot x _c$  completes the equilibrium of forces in $P_c$. The equation of motion in frontal plane and sagittal plane are independent.  By adding the external force (Disturbance) to the dynamics of the LIPM, the equations of motion can be modified and written as:
\begin{equation}
\begin{aligned}
\ddot x_{c}= \omega_n^{2}( x_{c}-p_{x}) +\frac{F_{ext,x}}{m}\hspace{1.1cm}\\
\ddot y_{c}= \omega_n^{2}( y_{c}-p_{y}) +\frac{F_{ext,y}}{m}\hspace{1.1cm}\\
\end{aligned}
 \label{eq:3}
\end{equation}

The effect of angular momentum of the upper-body, especially the torso and arms, can play an important role in push recovery. These joints can be used to apply a torque about the CoM. The CMP, is equal to the CoP in the case of zero torque about the CoM such as the LIPM. For a non-zero moment about the COM, however, the CMP can be out of the support polygon, while the COP still remains inside the support polygon. This effect can be embedded by considering the upper body as a flywheel that can be actuated directly as shown by Pratt \cite{pratt2006capture}.  In other words, the CMP is the point where a line parallel to the ground reaction force and passing through the COM intersects the ground. Therefore, by adding this effect to the LIPM dynamics, the equations of motion can be written as:

\begin{equation}
\begin{aligned}
\ddot x_{c}= \omega_n^{2}( x_{c}-p_{x}) -\frac{\dot H_{y}}{mz}+\frac{F_{ext,x}}{m}\hspace{1.1cm}\\
\ddot y_{c}= \omega_n^{2}( y_{c}-p_{y}) +\frac{\dot H_{x}}{mz}+\frac{F_{ext,y}}{m}\hspace{1.1cm}\\
\end{aligned}
 \label{eq4}
\end{equation}
where $\dot{H}$ is the rate of upper-body angular momentum that can be handled by the torque of arm and trunk joints. The relation between the ZMP and the CMP can be written as: \cite{khadiv2015optimal}:

\begin{equation}
\begin{aligned}
CMP_{x} =p_{x}+\frac{\dot H_{y}}{F_z}\hspace{1.1cm}\\
CMP_{y} =p_{y}-\frac{\dot H_{x}}{F_z}\hspace{1.1cm}\\
\end{aligned}
 \label{eq5}
\end{equation}

As a result, combining (\ref{eq4}) and (\ref{eq5}), we obtain:
\begin{equation}
\begin{aligned}
\ddot x_{c}= \omega_n^{2}( x_{c}-CMP_{x}) +\frac{F_{ext,x}}{m}\hspace{1.1cm}\\
\ddot y_{c}= \omega_n^{2}( y_{c}-CMP_{y}) +\frac{F_{ext,y}}{m}\hspace{1.1cm}\\
\end{aligned}
 \label{eq6}
\end{equation}

When the moment about the CoM is non-zero, such as when a disturbance is applied, the CMP and ZMP will diverge and CMP can leave the support polygon for controlling the CP, when the CP is outside of the support polygon.

\subsection{Capture Point Dynamics}

The unstable part of the LIPM dynamics has been called the CP and can be defined as follows \cite{pratt2006capture,koolen2012capturability,englsberger2015three}: 
\begin{equation}
\begin{aligned}
{\xi_x}={x_c}+\frac{{\dt x_c}}{{\omega_n}} \hspace{1.1cm}\\
{\xi_y}={y_c}+\frac{{\dt y_c}}{{\omega_n}} \hspace{1.1cm}\\
\end{aligned}
  \label{eq7}
\end{equation}
From (\ref{eq7}), the CoM dynamics is given by:

\begin{equation}
\begin{aligned}
{\dot x_c}={{\omega_n}} ({\xi}-{ x_c} )\hspace{1.1cm}\\
{\dot y_c}={{\omega_n}} ({\xi}-{ y_c} )\hspace{1.1cm}\\
\end{aligned}
  \label{eq8}
\end{equation}

By differentiating (\ref{eq8}) and substituting (\ref{eq6}) the CP dynamics is given by:

\begin{equation}
\begin{aligned}
\dt{\xi}_x=\omega_n({\xi_x}-{CMP}_{x})+\frac{F_{ext,x}}{m\omega_n}\hspace{1.1cm}\\
\dt{\xi}_y=\omega_n({\xi_y}-{CMP}_{y})+\frac{F_{ext,y}}{m\omega_n}\hspace{1.1cm}\\
\end{aligned}
  \label{eq9}
\end{equation}

\begin{figure}[]
\centering
 \includegraphics[scale=0.49, trim ={6.6cm 5.3cm 2.5cm 6.3cm},clip]{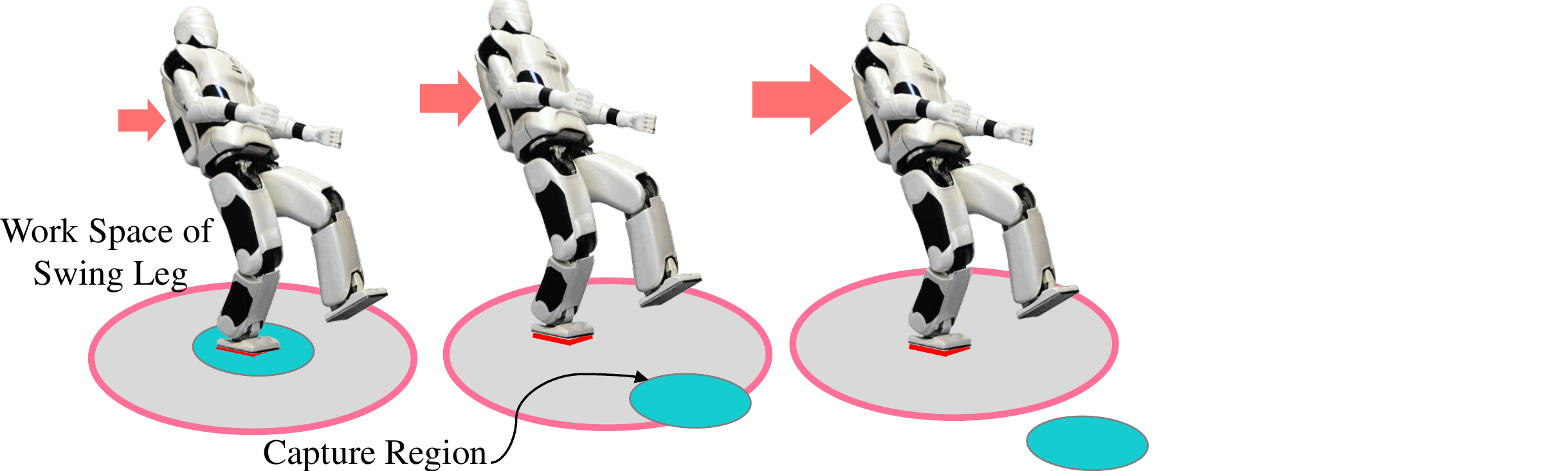}
\hfill\subcaptionbox{Ankle Strategy}[7em]{\centering}
 \subcaptionbox{Hip-ankle or stepping Strategy}[8em]{\centering } \subcaptionbox{Fail to push recovery
 in one step}[8em]{\centering }
  \hfill\null
\caption{Hip, ankle, and stepping strategy based on capture point\cite{pratt2006capture}}
      \label{fig2}
    \end{figure}

\begin{figure}[]
\centering  
 \includegraphics[scale=0.7, trim ={10.5cm 6.5cm 11cm 5.4cm},clip]{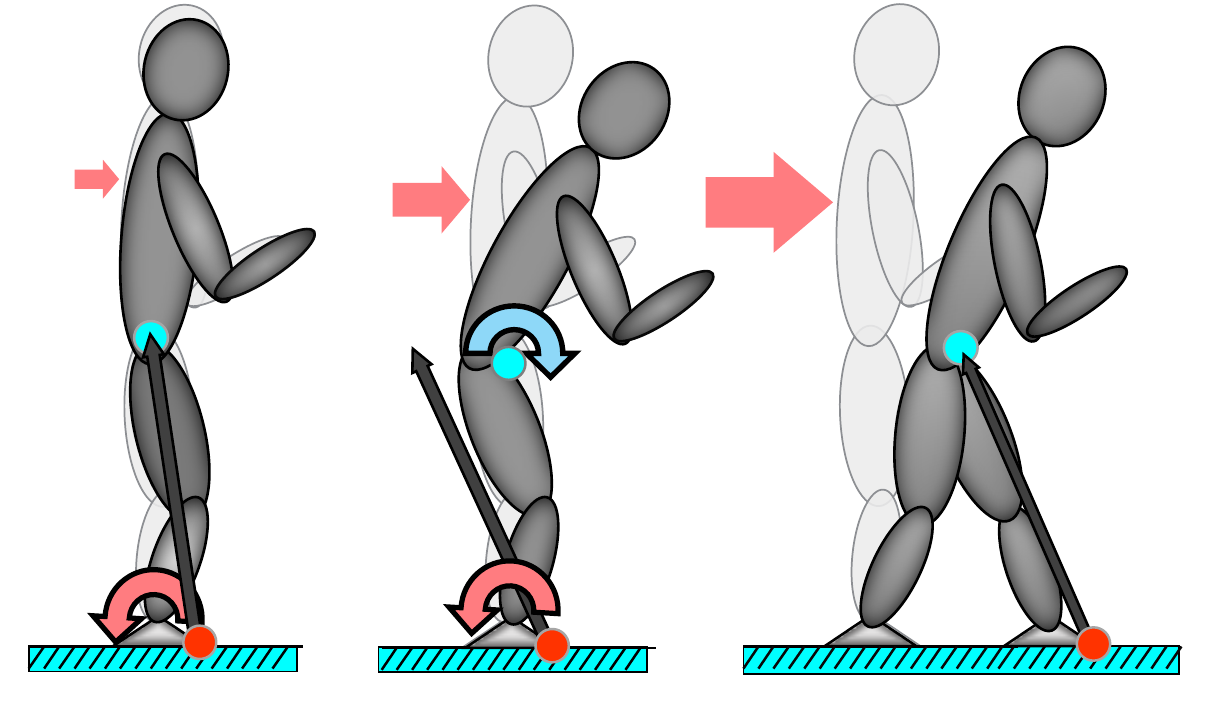}
\hfill\subcaptionbox{Ankle strategy}[7em]{\centering}
 \subcaptionbox{Hip strategy}[8em]{\centering } \subcaptionbox{Stepping strategy}[8em]{\centering }
  \hfill\null
\caption{Human-inspired balancing strategies}
      \label{fig3}
    \end{figure}

As it is obvious in (\ref{eq9}), the CMP can push the CP. In order to recover the balance of a humanoid robot, the CP should be controlled. When the CP is located within the support polygon, it can be controlled by the CoP \cite{englsberger2015three}, and when it is located out of the support polygon it can be controlled by the CMP or stepping.

Using the concept of CP we can determine when and where to take a step to recover from a push \cite{pratt2006capture}. If the CP is located within the support polygon, the robot is able to recover from the push without having to take a step. In order to stop in one step, the support polygon must have an intersection with the capture region as it shown on Fig.\ref{fig2}.(b),  \cite{pratt2006capture}. The robot will fail to recover from a severe push in one step, if the capture region does not intersect with the kinematic workspace of the swing foot. In the next sections we will discuss how to use the CP in Push recovery controller based on the MPC scheme.

\subsection{Human-Inspired Balancing Strategies}

\begin{figure}[]
\centering
 \includegraphics[scale=1.2, trim ={13.5cm 7.5cm 9cm 7.4cm},clip]{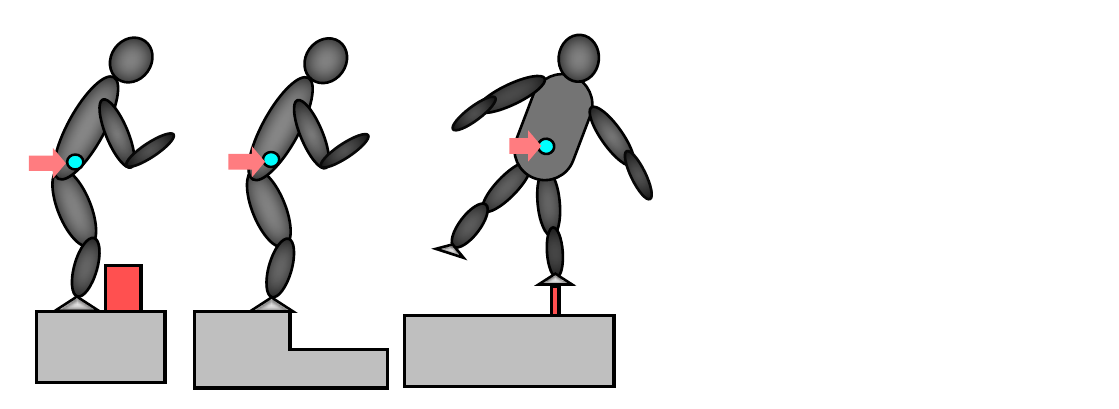}
\caption{Situations in which using stepping strategy is not possible}
      \label{fig4}
    \end{figure}

The response of a human to progressively increasing disturbances can be categorized into three basic strategy: (1) ankle strategy, (2) hip strategy (3) and stepping strategy. 
Humans tend to use the ankle strategy in case of small pushes to bring back the CP to its desired position as depicted in Fig.\ref{fig3}(a). However, the contact between the foot and floor is a unilateral constraint and if the ankle torque becomes too large, the CoP locates on the edge of the support polygon and the foot starts to rotate. Angular momentum of the upper body can be generated in the direction of the disturbance by applying a torque on the hip joint or arm joint as shown in Fig.\ref{fig3}(b).  This strategy also called CMP Balancing. With increasing the disturbance the useful strategy will be stepping Fig.\ref{fig3}(c).  However, there are several situations might occur where stepping is not possible as shown in Fig.\ref{fig4}. In this situation the balance recovery by Hip-Ankle strategy is necessary \cite{kiemel2012balance}.

Moreover in the situations that contact surface is small such as right side of  Fig.\ref{fig4}, generating upper body angular momentum for balance recovery is unavoidable. In this paper, the Hip-Ankle strategy is used in a single MPC scheme that will be presented in the following section.

\section {PUSH  RECOVERY CONTROLLER}
\subsection{discrete state-space form of LIPM+flywheel  dynamics}
We discretize the LIPM dynamics in the sagittal plane, while the procedure for the other direction is similar:
\begin{equation}
\begin{aligned}
 x_{c,t+1}= (1-\omega_n T) x_{c,t}+\omega_n T \xi_{x,t}\hspace{3.25cm}\\
  \xi_{x,t+1}= (1+\omega_n T)\xi_{x,t}-\omega_n T (p_{x,t}+\frac{\dot H_{y,t}}{mg}) +\frac{F_{ext,x}}{m\omega_n}\hspace{.79cm}\\
p_{x,t+1}= p_{x,t} +\dot p_{x,t}T\hspace{4.95cm}\\
\dot H_{y,t+1}= \dot H_{y,t} +\ddot H_{y,t}T\hspace{4.9cm}\\
\end{aligned}
 \label{eq5}
\end{equation}

This system can be re-written in discrete state-space form:

\begin{equation}
\mathbf{X}_{t+1}={A}_t \mathbf{X}_{t}+{B}\mathbf{U}_{t}
  \label{eq11}
\end{equation}
where $ \mathbf {X_t}=[x_{c,t},{\xi}_{x,t} ,p_{x,t}, \dot H_{y,t},  F_{ext,x}]  $ is the vector of state variables and $ U_t=[\dot p_{x,t},\ddot H_{y,t}]$ specifies the control inputs. The last state variable $ F_{EXT}$ is activated in the step time that a push is exerted by defining $\mu$. Therefore, when a push is exerted we have $\mu=1$ and in the other step times $\mu$ is equal to zero:
 \[ 
A_{t}=\begin{bmatrix}
  (1-\omega_n T)  & \omega_n T& 0  & 0 & 0\\[\verticaldistance]
0  & (1+\omega_n T)   & -\omega_n T &\frac{-\omega_n T}{mg} & \frac{1}{m\omega_n} \\[\verticaldistance]
  0 &0 &1  &0 & 0 \\[\verticaldistance]
    0 &0 &0  &1  & 0 \\[\verticaldistance]
       0 &0 &0  &0  & \mu 
\end{bmatrix}
\]

\[
B=
\begin{bmatrix}
    0 &0  \\
    0 & 0  \\
    T & 0  \\
    0 & T   \\
   0 & 0 
\end{bmatrix} 
\]

Given a sequence of control inputs  $\mathbf {\hat U}$, the linear model in (\ref{eq11}) can be converted into a sequence of states $\mathbf {\hat X}$, for the whole prediction horizon:

\begin{equation}
\begin{aligned}
\mathbf{\hat X}=\hat {\mathbf A}\mathbf{X}_{t}+\hat{\mathbf B}\mathbf{\hat U}\hspace{2cm}\\
\mathbf {\hat X}=[\mathbf {X^T_{t+1}}, \mathbf {X^T_{t+2}}, ..... , \mathbf { X^T_{t+N}}]\hspace{1cm}\\
\mathbf {\hat U}=[\mathbf {U^T_{t}}, \mathbf {U^T_{t+1}}, ..... , \mathbf {U^T_{t+N-1}}]\hspace{0.9cm}\\
\end{aligned}
  \label{eq12}
\end{equation}
where $\mathbf {\hat A}$  and $\mathbf {\hat B}$  are defined recursively from (\ref{eq11}). The control inputs are the rate of change of ZMP position and the rate of upper-body angular momentum. As a result, the core of the Proposed MPC is based on combined hip and ankle strategies.

\subsection{Model Predictive Control (MPC)}
We present an MPC Controller that uses the concept of hip and ankle strategies in its core by modulating the ZMP and CMP as control inputs, considering future constraints on the CP . Using the LIPM+Flywheel, the trajectory optimization is simplified to a Quadratic Programming (QP) problem. The LIPM+Flywheel has a linear dynamics and the corresponding optimization problem is linear and can be solved in real-time.
The push recovery control objective is simplified to optimize control inputs subject to terminal constraints on the CP, CMP and change of angular momentum. Constraints will be discussed in the next subsection. The objective function used in this paper is as follows:
\begin{equation}
\begin{aligned}
J= \sum_{\mathclap{k=1}}^{N}  \alpha_{1}\|  \xi_{k+1}-\xi^{ref}_{k+1}    \|_{x}^2  +\alpha_{2}\|  \ddt H_{k}    \|_{x}^2+ \alpha_{3}\|  \dt {cop}_{k}   \|_{x}^2 +  \alpha_{4}\|   \dt H_{k}   \|_{y}^2\\ +\alpha_{5}\|  \xi_{k+1}^{ref}- \xi_{k+1}   \|_{y}^2 +\alpha_{6}\|  \ddt H_{k}    \|_{y}^2+\hspace{0.2cm} 
\alpha_{7}\|  \dt {cop}_{k}   \|_{y}^2 +
 \alpha_{8}\|   \dt H_{k}   \|_{x}^2
  \hspace{0.01cm}
\\[10pt]
\end{aligned}
 \label{eq13} 
\end{equation}
where  $ \dot P_x ,\dot P_y, \ddot H_y$ and $ \ddot H_x$ are vectors of control inputs over the next $ N$ time steps. The first term minimizes distance between the desired and actual CP. The second and third terms are considered for modulating the ZMP and CMP in order to control the CP. The forth term is used for minimizing the rate of change of angular momentum. The $\alpha_i$ are the weights each term that can be regulated in different situations. This proposed cost function consider both rotational and linear dynamics of biped robots. The proposed objective function can be converted to the following standard quadratic form:
\begin{equation}
\begin{aligned}
J=\frac{1}2 \hspace{.1cm}\mathbf {\hat U}^TH\hspace{0.1cm}\mathbf {\hat U} +\hspace{.1cm} \mathbf {\hat U}^T f\\
st. \hspace{1.7cm}\\
C \hspace{.1cm} \mathbf {\hat U}+D=0 \hspace{1.1cm}\\
E\hspace{.1cm} \mathbf {\hat U}+F \leq0  \hspace{1.1cm}
\end{aligned}
  \label{eq14}
\end{equation}
where $A, B, C$ and $D$ are coefficient matrices, with $H$ and $f$ being the Hessian matrix and gradient vector of the objective function respectively.

\subsection{Constraints}
The real power of MPC is the consideration of future constraints.  Our goal in push recovery controller is to maintain the ZMP inside the support polygon and controlling the CP by modulating the ZMP and CMP. Furthermore, we need to coincide the CP with the ZMP in the support polygon center at the end of motion, while the rate of change of upper-body angular momentum is zero. This means we have the following constraints:
\begin{equation}
\begin{aligned}
{\xi}_{x,N}={\xi_{ref,x}}\hspace{1.1cm}\\
{x}_{c,N}={\xi_{ref,x}}\hspace{1.1cm}\\
{p}_{x,N}={\xi_{ref,x}}\hspace{1.1cm}\\
{\dt{H}}_{y,N}=0 \hspace{1.1cm}\\
 {p}_{x ,i}  \in   Support Polygon 
\end{aligned}
  \label{eq15}
\end{equation}
where ${\xi_{ref,x}}$ is the reference CP that is located on the center of support polygon. The first four constraints are equality constraints for the last step time of motion. The last equation is an inequality constraint that enforces the ZMP to remain inside the support polygon. Similar equations can be derived for the lateral direction. Using the objective function of (\ref{eq13}) and adding the constraints of (\ref{eq15}), control inputs can be optimized during push recovery by the QP.

\section {SIMULATION AND DISCUSSION}
To verify the performance of the push recovery controller, we performed simulations using MATLAB. The proposed controller is implemented on an abstract model of the SURENA \rom{3} humanoid robot. Parameters that have been used in the simulation is shown in  Table.\ref{Tab1} The time of balance recovery is considered 1.5 s. The allowable rate of upper-body angular momentum that can be applied is 190 N.m during 1.5 s according to \cite{aftab2012ankle}.

\begin{table}[t]
\caption{Variable used in the simulation(Based on SURENA \rom{3})}
\label{table_example}
\begin{center}
\begin{tabular}{@{\hskip 0.35in} c @{\hskip 0.5in} c @{\hskip 0.5in} c  @{\hskip 0.3in}}
\toprule
Variable & Symbol  & Value \\
\midrule
Height & - &190 cm\\
CoM Height & $z_c$ &   75 cm\\
 Mass & m & 98 kg \\
Foot Length & -  &   25 cm\\
Foot Width& -& 15 cm\\
Trunk Inertia &-  &$ 8 \, \, \text{Kg.m}^2$ \\
Arms Inertia &  - & $ 3 \, \, \text{Kg.m}^2$ \\
Step-time  & T & $ 0.05 \,\, \text{s}$ \\
MPC gain &$\alpha_{1}$    & $1 \, \, {\text{m}^{-1}}$ \\
MPC gain &$\alpha_{2}$    &   $3\, \,  {s}.{(\text{N.m})}^{-1}$ \\
MPC gain&$\alpha_{3}$   & $10^{-6} \, \,  {\text{s}}.{\text{m}^{-1}}$\\
MPC gain&$\alpha_{4}$  & $10^{-3}  \, \, {(\text{N.m})}^{-1}$ \\
MPC gain&$\alpha_{5}$   & $1\, \, {\text{m}^{-1}}$ \\
MPC gain&$\alpha_{6}$   & $1\, \, {\text{s}}.{(\text{N.m})}^{-1}$\\
MPC gain&$\alpha_{7}$    & $10^{-6} \, \,  {\text{s}}.{\text{m}^{-1}}$\\
MPC gain&$\alpha_{8}$   & $10^{-3} \, \, {(\text{N.m})}^{-1}$\\
\bottomrule
\end{tabular}
\end{center}
  \label{Tab1}
\end{table}

\subsection{Simulation Results}
\begin{figure}
    \includegraphics[,scale=0.550, trim ={3.5cm 11.2cm 0.5cm 2.5cm},clip]{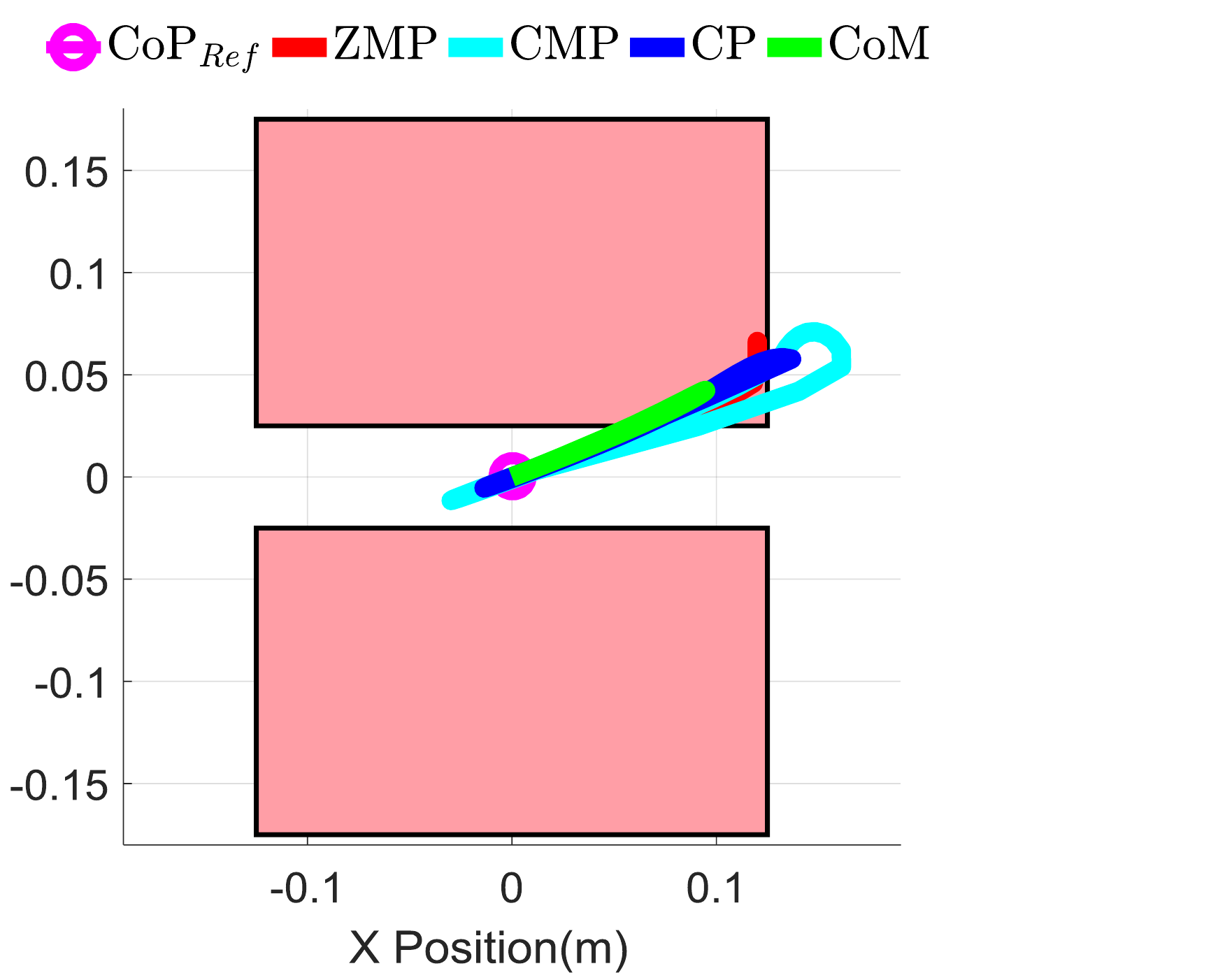}\par
 \includegraphics[,scale=0.53, trim ={2.61cm 13.3cm 2.5cm 2cm},clip]{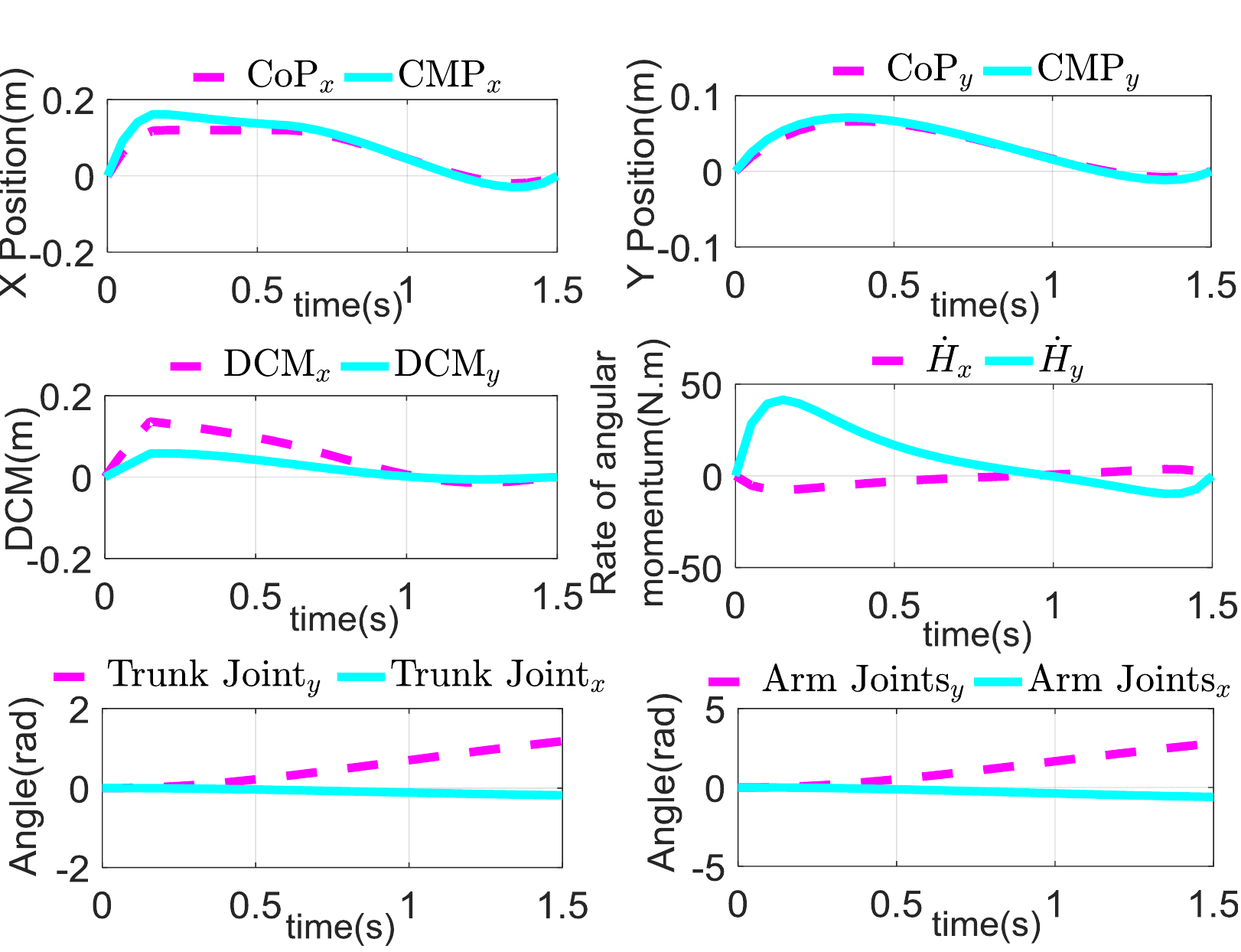}\par
\caption{Simulation results of push recovery controller, the push with magnitude of (360,140) N is exerted on the CoM (The robot stands on both legs)}
 \label{fig5}
\end{figure}

In the first scenario, a push with the magnitude of 360 N in sagittal direction, and another one with the magnitude of 140 N in frontal plane are exerted on the CoM of the robot. As we expected, the large push throws the CP out of the support polygon, and the ZMP cannot navigate it. Therefore, the angular momentum is generated by the MPC to move the CMP outside the support polygon for controlling the CP. The maximum flywheel torque for push recovery is about 50 N.m that is realizable on our considered robot. The trajectory of CP, CoP, CMP and CoM during balance recovery is shown in  Fig.\ref{fig5}. 

\begin{figure*}
\begin{multicols}{2}
 \includegraphics[,scale=0.55, trim ={3.cm 13.5cm 3cm 2.3cm},clip]{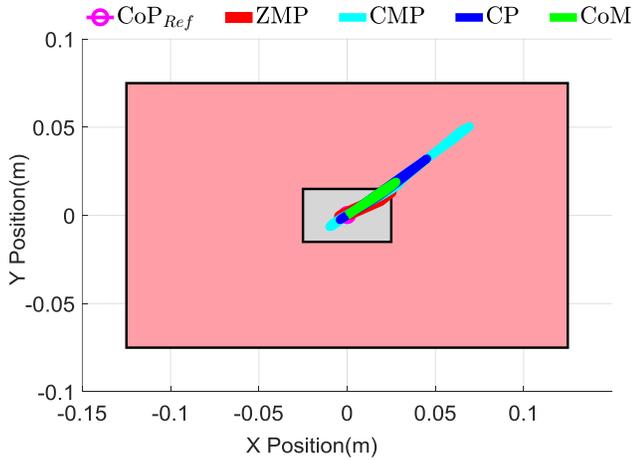}\par 
    \includegraphics[,scale=0.55, trim ={2.5cm 14cm 3cm 2.3cm},clip]{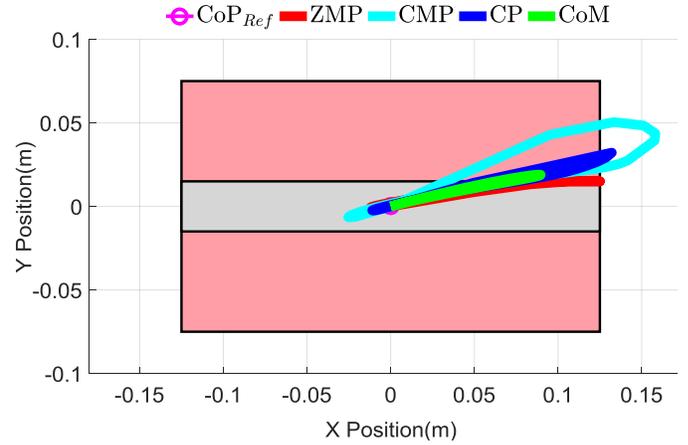}\par 
    \end{multicols}
\begin{multicols}{2}
    \includegraphics[,scale=0.43, trim ={0.5cm 10cm 0.5cm 3cm},clip]{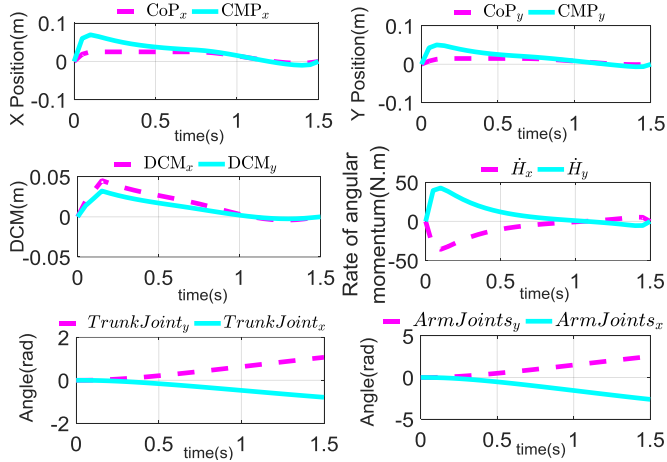}\par
\subcaption{The push (140,100) N is exerted during 0.05 s, while the contact surface is a point}
 \includegraphics[,scale=0.43, trim ={0.5cm 9cm 0.4cm 4cm},clip]{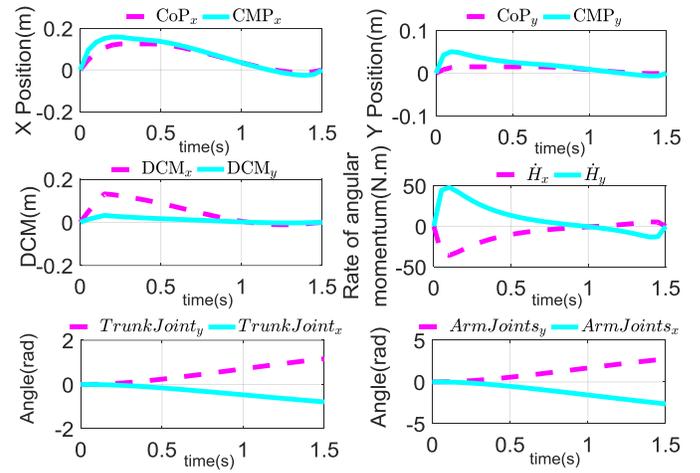}\par
\subcaption{The push (350,100) N is exerted during 0.05 s, while the contact surface is a line}
\end{multicols}
\caption{Simulation results of push recovery (The robot stands on one leg)}
 \label{fig6}
\end{figure*}

In the second scenario, the robot stands on one leg, while the contact surface is shrunken to a line or a point. Two examples for this situation are standing on lumber and standing on rock. In this case, the CMP modulation recovers the robot from the disturbance, because the support polygon is too small and the ankle strategy is not helpful anymore. In this situation, the CP leaves the support polygon and the CoP remains on the bounds of the support polygon, while the CMP pushes the CP to the desired position. As shown in  Fig.\ref{fig6}, in the first case, pushes with magnitude of 350 N in sagittal and 100 N in lateral direction are exerted on the CoM, while the surface contact is a line. In the second case, pushes with magnitude of 140 N in sagittal and 100 N in frontal plane are exerted on the CoM, while the surface contact is a point\footnote{A summary of the simulation scenarios is available on https://youtu.be/bDPafm-6CLk}. 

As shown in  Fig.\ref{fig5}, \ref{fig6}, in all simulations the angle of the hip pitch joint is smaller than 1.5 rad that is allowable \cite {aftab2012ankle}.

Based on the simulation results, the regulation of angular momentum is so beneficial during push recovery, especially in the standing on small contact surfaces or in the situations where stepping is not possible. Based on presented results the proposed method has the following features:

\begin{itemize}
\item The presented MPC scheme is capable of generating human-like response to external disturbances; for example, when the exerted force is small, it uses the ankle strategy for balance recovery. Furthermore, in the presence of large disturbances, it generates angular momentum and uses hip-ankle strategy simultaneously.

\item The proposed push recovery controller can compensate the severe pushes, when the robot stands on small contacts such as a line or a point and also is capable of saving the robot from falling in the situations that stepping is not possible.
\end{itemize}

\section{CONCLUSION AND FUTURE WORK}
In this paper, a push recovery controller based on the CP concept and through an MPC framework is developed. The core of the proposed MPC is based on a combined hip and ankle strategies by modulating the CMP and ZMP to control the CP. The results showed that this controller is capable of rejecting severe pushes, even in the case where the support polygon is limited to a line or a point, and stepping is not allowed. The effectiveness of the proposed MPC scheme was demonstrated by simulating an abstract model of the SURENA III humanoid robot. 

Despite all above advantages, this controller is implemented only in simulation. Implementing on the experimental setup has more practical challenges\cite{wiedebach2016walking}. For example, accurate state estimation to obtain the CP position, the saturation of actuators especially in the case where the support polygon is a point or a line, foot slipping and bringing the upper-body back into an upright position are some of main challenges of experimental implementation that will be discussed in the future works.





\bibliographystyle{IEEEtran}
\bibliography{Master}

\begin{thebibliography}{10}
\providecommand{\url}[1]{#1}
\csname url@samestyle\endcsname
\providecommand{\newblock}{\relax}
\providecommand{\bibinfo}[2]{#2}
\providecommand{\BIBentrySTDinterwordspacing}{\spaceskip=0pt\relax}
\providecommand{\BIBentryALTinterwordstretchfactor}{4}
\providecommand{\BIBentryALTinterwordspacing}{\spaceskip=\fontdimen2\font plus
\BIBentryALTinterwordstretchfactor\fontdimen3\font minus
  \fontdimen4\font\relax}
\providecommand{\BIBforeignlanguage}[2]{{%
\expandafter\ifx\csname l@#1\endcsname\relax
\typeout{** WARNING: IEEEtran.bst: No hyphenation pattern has been}%
\typeout{** loaded for the language `#1'. Using the pattern for}%
\typeout{** the default language instead.}%
\else
\language=\csname l@#1\endcsname
\fi
#2}}
\providecommand{\BIBdecl}{\relax}
\BIBdecl

\bibitem{kajita2003biped}
S.~Kajita, F.~Kanehiro, K.~Kaneko, K.~Fujiwara, K.~Harada, K.~Yokoi, and
  H.~Hirukawa, ``Biped walking pattern generation by using preview control of
  zero-moment point,'' in \emph{Robotics and Automation, 2003. Proceedings.
  ICRA'03. IEEE International Conference on}, vol.~2.\hskip 1em plus 0.5em
  minus 0.4em\relax IEEE, 2003, pp. 1620--1626.

\bibitem{wieber2006trajectory}
P.-B. Wieber, ``Trajectory free linear model predictive control for stable
  walking in the presence of strong perturbations,'' in \emph{2006 6th IEEE-RAS
  International Conference on Humanoid Robots}.\hskip 1em plus 0.5em minus
  0.4em\relax IEEE, 2006, pp. 137--142.

\bibitem{pratt2006capture}
J.~Pratt, J.~Carff, S.~Drakunov, and A.~Goswami, ``Capture point: A step toward
  humanoid push recovery,'' in \emph{2006 6th IEEE-RAS international conference
  on humanoid robots}.\hskip 1em plus 0.5em minus 0.4em\relax IEEE, 2006, pp.
  200--207.

\bibitem{herdt2010online}
A.~Herdt, H.~Diedam, P.-B. Wieber, D.~Dimitrov, K.~Mombaur, and M.~Diehl,
  ``Online walking motion generation with automatic footstep placement,''
  \emph{Advanced Robotics}, vol.~24, no. 5-6, pp. 719--737, 2010.

\bibitem{stephens2010push}
B.~J. Stephens and C.~G. Atkeson, ``Push recovery by stepping for humanoid
  robots with force controlled joints,'' in \emph{2010 10th IEEE-RAS
  International Conference on Humanoid Robots}.\hskip 1em plus 0.5em minus
  0.4em\relax IEEE, 2010, pp. 52--59.

\bibitem{aftab2012ankle}
Z.~Aftab, T.~Robert, and P.-B. Wieber, ``Ankle, hip and stepping strategies for
  humanoid balance recovery with a single model predictive control scheme,'' in
  \emph{2012 12th IEEE-RAS International Conference on Humanoid Robots
  (Humanoids 2012)}.\hskip 1em plus 0.5em minus 0.4em\relax IEEE, 2012, pp.
  159--164.

\bibitem{koolen2012capturability}
T.~Koolen, T.~De~Boer, J.~Rebula, A.~Goswami, and J.~Pratt,
  ``Capturability-based analysis and control of legged locomotion, part 1:
  Theory and application to three simple gait models,'' \emph{The International
  Journal of Robotics Research}, vol.~31, no.~9, pp. 1094--1113, 2012.

\bibitem{englsberger2015three}
J.~Englsberger, C.~Ott, and A.~Albu-Sch{\"a}ffer, ``Three-dimensional bipedal
  walking control based on divergent component of motion,'' \emph{IEEE
  Transactions on Robotics}, vol.~31, no.~2, pp. 355--368, 2015.

\bibitem{krause2012stabilization}
M.~Krause, J.~Englsberger, P.-B. Wieber, and C.~Ott, ``Stabilization of the
  capture point dynamics for bipedal walking based on model predictive
  control,'' \emph{IFAC Proceedings Volumes}, vol.~45, no.~22, pp. 165--171,
  2012.

\bibitem{yun2011momentum}
S.-k. Yun and A.~Goswami, ``Momentum-based reactive stepping controller on
  level and non-level ground for humanoid robot push recovery,'' in \emph{2011
  IEEE/RSJ International Conference on Intelligent Robots and Systems}.\hskip
  1em plus 0.5em minus 0.4em\relax IEEE, 2011, pp. 3943--3950.

\bibitem{kiemel2012balance}
S.~Kiemel, ``Balance maintenance of a humanoid robot using the hip-ankle
  strategy,'' Ph.D. dissertation, TU Delft, Delft University of Technology,
  2012.

\bibitem{wiedebach2016walking}
G.~Wiedebach, S.~Bertrand, T.~Wu, L.~Fiorio, S.~McCrory, R.~Griffin, F.~Nori,
  and J.~Pratt, ``Walking on partial footholds including line contacts with the
  humanoid robot atlas,'' \emph{arXiv preprint arXiv:1607.08089}, 2016.

\bibitem{khadiv2015optimal}
M.~Khadiv, S.~A.~A. Moosavian, A.~Yousefi-Koma, M.~Sadedel, and S.~Mansouri,
  ``Optimal gait planning for humanoids with 3d structure walking on slippery
  surfaces,'' \emph{Robotica}, vol.~3, pp. 1--19, 2015.

\bibitem{kajita20013d}
S.~Kajita, F.~Kanehiro, K.~Kaneko, K.~Yokoi, and H.~Hirukawa, ``The 3d linear
  inverted pendulum mode: A simple modeling for a biped walking pattern
  generation,'' in \emph{Intelligent Robots and Systems, 2001. Proceedings.
  2001 IEEE/RSJ International Conference on}, vol.~1.\hskip 1em plus 0.5em
  minus 0.4em\relax IEEE, 2001, pp. 239--246.

\bibitem{popovic2005ground}
M.~B. Popovic, A.~Goswami, and H.~Herr, ``Ground reference points in legged
  locomotion: Definitions, biological trajectories and control implications,''
  \emph{The International Journal of Robotics Research}, vol.~24, no.~12, pp.
  1013--1032, 2005.

\end{thebibliography}

\end{document}